\documentclass[11pt,a4paper]{article}
\usepackage[hyperref]{acl2021}
\usepackage{times}
\usepackage{latexsym}

\usepackage{microtype}

\usepackage{amsmath}
\usepackage{amsthm}
\usepackage{amssymb}
\usepackage{multirow}
\usepackage{xcolor}
\usepackage{lipsum}
\usepackage{comment}
\usepackage{booktabs}
\usepackage{subcaption} %To add multiple images in one image
\usepackage{graphicx}

\aclfinalcopy % Uncomment this line for the final submission
\theoremstyle{definition}
\newtheorem{definition}{Definition}[section]
\newcommand{\norm}[1]{\left\lVert#1\right\rVert}
\newcommand{\abs}[1]{\mid#1\mid}
\renewcommand*{\vec}[1]{\mathbf{#1}}
\newcommand{\mat}[1]{\mathbf{#1}}

\title{Poisoning Knowledge Graph Embeddings via Relation Inference Patterns}

\author{Peru Bhardwaj$^{1}$ \quad John Kelleher$^{2}$\thanks{\ \ Equal contribution by last authors.} \quad Luca Costabello$^{3*}$ \quad Declan O'Sullivan$^{1*}$  \\ 
	$^1$ ADAPT Centre, Trinity College Dublin, Ireland \\
	$^2$ ADAPT Centre, TU Dublin, Ireland \\
	$^3$ Accenture Labs, Ireland \\
	\texttt{peru.bhardwaj@adaptcentre.ie} 
}

\date{}

\begin{document}
\maketitle
\begin{abstract}
    We study the problem of generating data poisoning attacks against Knowledge Graph Embedding (KGE) models for the task of link prediction in knowledge graphs. To poison KGE models, we propose to exploit their inductive abilities which are captured through the relationship patterns like symmetry, inversion and composition in the knowledge graph. Specifically, to degrade the model's prediction confidence on target facts, we propose to improve the model's prediction confidence on a set of decoy facts. Thus, we craft adversarial additions that can improve the model's prediction confidence on decoy facts through different inference patterns. Our experiments demonstrate that the proposed poisoning attacks outperform state-of-art baselines on four KGE models for two publicly available datasets. We also find that the symmetry pattern based attacks generalize across all model-dataset combinations which indicates the sensitivity of KGE models to this pattern. 
\end{abstract}

% -------------------------------------------------------------------------------------
% -------------------------------------------------------------------------------------

\section{Introduction}
Knowledge graph embeddings (KGE) are increasingly deployed in domains with high stake decision making like healthcare and finance \citep{noy2019knowledgegraphs}, where it is critical to identify the potential security vulnerabilities that might cause failure. But the research on adversarial vulnerabilities of KGE models has received little attention. 
We study the adversarial vulnerabilities of KGE models through data poisoning attacks. 
These attacks craft input perturbations at training time that aim to subvert the learned model's predictions at test time. 

Poisoning attacks have been proposed for models that learn from other graph modalities \citep{xu2020advgraphsurvey} but they cannot be applied directly to KGE models. This is because they rely on  gradients of all possible entries in a dense adjacency matrix and thus, do not scale to large knowledge graphs with multiple relations.
The main challenge in designing poisoning attacks for KGE models is the large combinatorial search space of candidate perturbations which is of the order of millions for benchmark knowledge graphs with thousands of nodes. Two recent studies \cite{zhang2019kgeattack, pezeshkpour2019criage} attempt to address this problem through random sampling of candidate perturbations \citep{zhang2019kgeattack} or through a vanilla auto-encoder that reconstructs discrete entities and relations from latent space \cite{pezeshkpour2019criage}. 
However, random sampling depends on the number of candidates being sampled and the auto-encoder proposed in \citet{pezeshkpour2019criage} is only applicable to multiplicative KGE models.

\captionsetup[figure]{font=small}
\begin{figure}[]
    \centering
    \begin{subfigure}[htb]{1\columnwidth}
        \includegraphics[width=1\columnwidth]{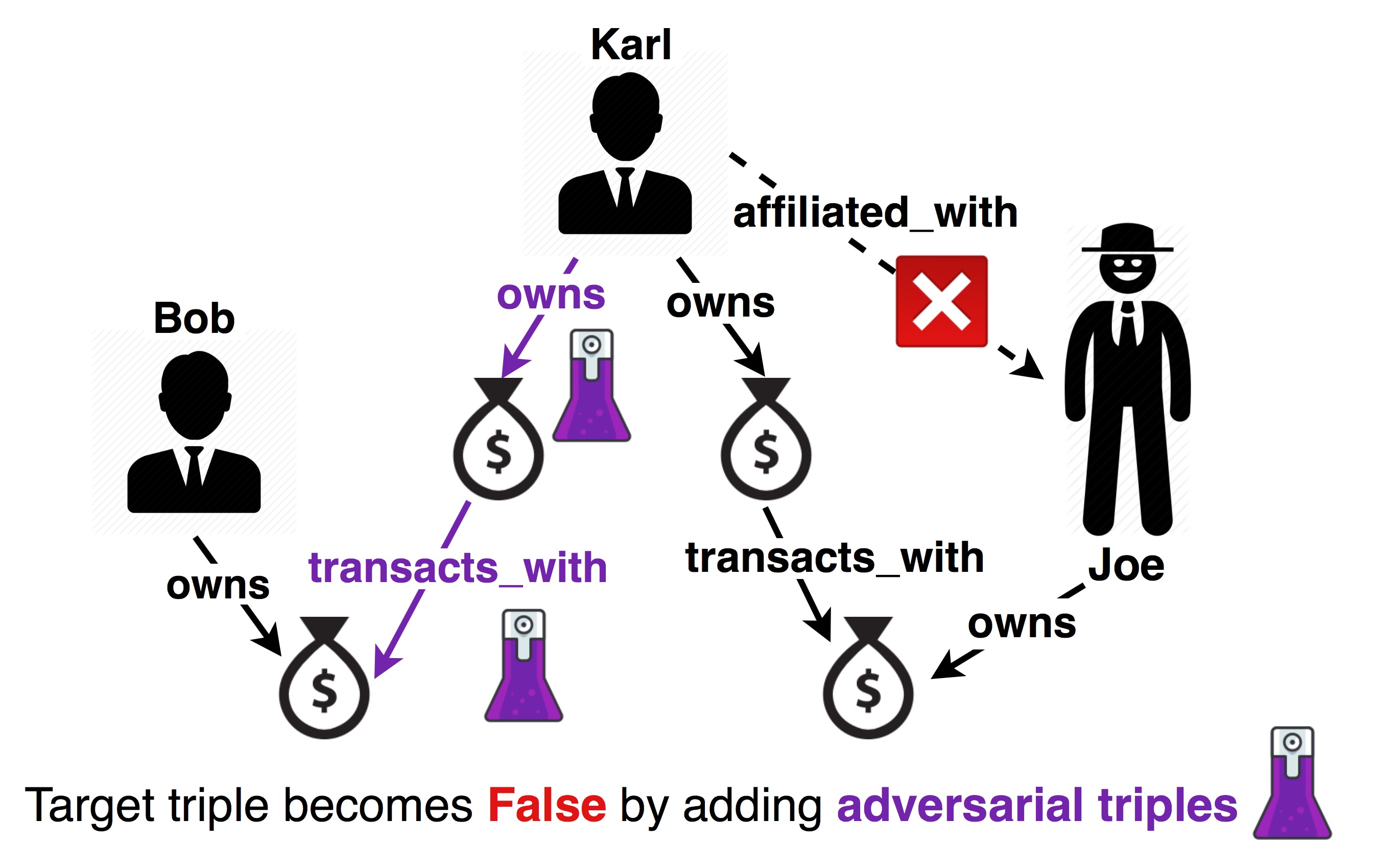}
        
    \end{subfigure}
    
    \caption{Composition based adversarial attack on fraud detection. The knowledge graph consists of two types of entities - Person and BankAccount. The target triple to predict is $(Karl, \mathtt{affiliated\_with}, Joe\_the\_mobster)$. Original KGE model predicts this triple as True. But a malicious attacker adds adversarial triples (in purple) that connect $Karl$ with a non-suspicious person $Bob$ through composition pattern. Now, the KGE model predicts the target triple as False.
    }
    \label{fig:example}
\end{figure}

In this work, we propose to exploit the inductive abilities of KGE models to craft poisoned examples against the model. 
The inductive abilities of KGE models are expressed through different connectivity patterns like symmetry, inversion and composition between relations in the knowledge graph.
We refer to these as \emph{inference patterns}.
We focus on the task of link prediction using KGE models and consider the adversarial goal of \emph{degrading} the predicted rank of \emph{target} missing facts. 
To degrade the ranks of target facts, we propose to carefully select a set of decoy facts and exploit the inference patterns to improve performance on this decoy set.
Figure \ref{fig:example} shows an example of the use of composition pattern to degrade KGE model's performance.

We explore a collection of heuristic approaches to select the decoy triples and craft adversarial perturbations that use different inference patterns to improve the model's predictive performance on these decoy triples. Our solution addresses the challenge of large candidate space by breaking down the search space into smaller steps - (i) determining adversarial relations; (ii) determining the decoy entities that most likely violate an inference pattern; and (iii) determining remaining adversarial entities in the inference pattern that are most likely to improve the rank of decoy triples. 

We evaluate the proposed attacks on four state-of-art KGE models with varied inductive abilities - DistMult, ComplEx, ConvE and TransE. We use two publicly available benchmark datasets for link prediction - WN18RR and FB15k-237. Comparison against the state-of-art poisoning attacks for KGE models shows that our proposed attacks outperform them in \emph{all} cases. We find that the attacks based on symmetry pattern perform the best and generalize across all model-dataset combinations.

Thus, the main contribution of our research is an effective method to generate data poisoning attacks, which is based on inference patterns captured by KGE models. Through a novel reformulation of the problem of poisoning KGE models, we overcome the existing challenge in the scalability of poisoning attacks for KGE models. 
Furthermore, the extent of effectiveness of the attack relying on an inference pattern indicates the KGE model's sensitivity to that pattern. 
Thus, our proposed poisoning attacks help in understanding the KGE models.

% -------------------------------------------------------------------------------------------------------
% -------------------------------------------------------------------------------------------------------

\section{Problem Formulation}
For a set of entities \( \mathcal{E} \) and a set of relations \( \mathcal{R} \), a knowledge graph is a collection of triples represented as \(
 \mathcal{KG} = \{(s,\mathtt{r},o)\, |\,  s,o \in \mathcal{E} \, and \, \mathtt{r} \in \mathcal{R} \}
  \), where $s, \mathtt{r}, o$ represent the subject, relation and object in a triple.
A Knowledge Graph Embedding (KGE) model encodes entities and relations to a low-dimensional continuous vector space \( \vec{e}_s, \vec{e}_\mathtt{r}, \vec{e}_o \in \mathbb{R}^{k} \) where \( k \) is the embedding dimension. To do so, it uses a scoring function \(f : \mathcal{E} \times \mathcal{R} \times \mathcal{E} \rightarrow \mathbb{R} \) which depends on the entity and relation embeddings to assign a score to each triple \(f_{sro} = f(\vec{e}_s, \vec{e}_\mathtt{r}, \vec{e}_o) \). Table \ref{tab:scoring_functions} shows the scoring functions of state-of-art KGE models studied in this research. The embeddings are learned such that the scores for true (existing) triples in the knowledge graph are higher than the scores for false (non-existing) triples in the knowledge graph. 

\begin{table}[]
    \centering
    \small
    \begin{tabular}{|c|c|}
    \hline
    \textbf{Model} &  \textbf{Scoring Function}\\
    \hline
    \hline
        
         DistMult  & $ \langle \vec{e}_s, \vec{e}_\mathtt{r}, \vec{e}_o \rangle$ \\
         \hline
         ComplEx  & $ \Re(\langle \vec{e}_s, \vec{e}_\mathtt{r}, \overline{\vec{e}_o} \rangle)$  \\
         \hline
         ConvE  & $ \langle \sigma(\mathrm{vec}(\sigma([ \overline{\vec{e}_\mathtt{r}}, \overline{\vec{e}_s}] \ast \boldsymbol{\Omega})) \mat{W}), \vec{e}_o \rangle$ \\
         \hline
         TransE  &  $-\norm{\vec{e}_s + \vec{e}_\mathtt{r} - \vec{e}_o}$    \\
        \hline
         
    \end{tabular}
    \caption{Scoring functions \(f_{sro}\) of the KGE models used in this research. For ComplEx, $\vec{e}_s, \vec{e}_\mathtt{r}, \vec{e}_o \in \mathbb{C}^k$; for the remaining models $\vec{e}_s, \vec{e}_\mathtt{r}, \vec{e}_o \in \mathbb{R}^k$. Here, $\langle \cdot \rangle$ denotes the tri-linear dot product; $\sigma$ denotes sigmoid activation function, $\ast$ denotes 2D convolution; $\overline{\ \cdot\ }$ denotes conjugate for complex vectors, and 2D reshaping for real vectors in ConvE model; $\norm{\cdot}$ denotes l-p norm}
    \label{tab:scoring_functions}
\end{table}

\textbf{Multiplicative vs Additive Interactions: }
The scoring functions of KGE models exhibit multiplicative or additive interactions \cite{chandrahas2018towards}. The multiplicative models score triples through multiplicative interactions of subject, relation and object embeddings. The scoring function for these models can be expressed as $f_{sro} = \vec{e}_{\mathtt{r}}^\top \mathcal{F}(\vec{e}_s, \vec{e}_o)$ where the function $\mathcal{F}$ measures the compatibility between the subject and object embeddings and varies across different models within this family. DistMult, ComplEx and ConvE have such interactions. On the other hand, additive models score triples through additive interactions of subject, relation and object embeddings. The scoring function for such models can be expressed as $f_{sro} = -\norm{\mat{M}_{\mathtt{r}}^1(\vec{e}_s) + \vec{e}_\mathtt{r} - \mat{M}_{\mathtt{r}}^2(\vec{e}_o)}$ where \( \vec{e}_s, \vec{e}_o \in \mathbb{R}^{k_\mathcal{E}} \), \(\vec{e}_\mathtt{r} \in \mathbb{R}^{k_\mathcal{R}} \) and $\mat{M}_\mathtt{r} \in \mathbb{R}^{k_{\mathcal{E}} \times k_{\mathcal{R}}}$ is the projection matrix from entity space $\mathbb{R}^{k_{\mathcal{E}}}$ to relation space $\mathbb{R}^{k_{\mathcal{R}}}$. TransE has additive interactions. 

\textbf{Inductive Capacity of KGE models:}
The general intuition behind the design of the scoring functions of KGE models is to capture logical properties between relations from the observed facts in the knowledge graph. These logical properties or \emph{inference patterns} can then be used to make downstream inferences about entities and relations. For example, the relation $ \mathtt{is\_owned\_by}$ is inverse of the relation $ \mathtt {owns}$, and when the fact $(Account42,\mathtt{is\_owned\_by}, Karl)$ is true, then the fact $(Karl,\mathtt {owns}, Account42)$ is also true and vice versa. A model that can capture inversion pattern can thus predict missing facts about $\mathtt {owns}$ based on observed facts about $\mathtt{is\_owned\_by}$. The most studied inference patterns in the current literature are symmetry, inversion and composition since they occur very frequently in real-world knowledge graphs. In this work, we use these patterns to investigate the adversarial vulnerability of KGE models.

\textbf{Link Prediction:}
Since most of the existing knowledge graphs are incomplete, a standard use case of KGE models is to predict missing triples in the \( \mathcal{KG}\). 
This task is evaluated by an entity ranking procedure. Given a test triple $(s,\mathtt{r},o)$, the subject entity is replaced by each entity from \( \mathcal{E} \) in turn. These replacements are referred to as \emph{synthetic negatives}. 
The KGE model's scoring function is used to predict scores of these negative triples. The scores are then sorted in descending order and the rank of the correct entity is determined. These steps are repeated for the object entity of the triple. 

The state-of-art evaluation metrics for this task are (i) \emph{MR} which is the mean of the predicted ranks, (ii) \emph{MRR} which is the mean of the reciprocals of predicted ranks and (iii) \emph{Hits@n} which count the proportion of correct entities ranked in top-n. 
In the filtered setting \citep{bordes2013transe}, negative triples that already exist in the training, validation or test set are filtered out. That is, their scores are ignored while computing the ranks. Depending on the domain of use, either subject or object or both ranks of the test triple are used to determine the model's confidence\footnote{KGE models do not provide model uncertainty estimates.} in predicting a missing link.

\paragraph{Poisoning Attacks on KGE models: }
We study poisoning attacks for the task of link prediction using KGE models. We focus on targeted attacks where the attacker targets a specific set of missing triples instead of the overall model performance. We use the notation $(s,\mathtt{r},o)$ for the \emph{target triple}; in this case, $s,o$ are the \emph{target entities} and $\mathtt{r}$ is the \emph{target relation}. The goal of an adversarial attacker is to degrade the ranks of missing triples which are predicted highly plausible by the model. 
The rank of a highly plausible target triple can be degraded by improving the rank of less plausible \emph{decoy triples}. For a target triple $(s,\mathtt{r},o)$, the decoy triple for degrading the rank on object side would be $(s,\mathtt{r},o')$ and the decoy triple for degrading the rank on subject side would be $(s',\mathtt{r},o)$. 
Thus, the aim of the adversarial attacker is to select decoy triples from the set of valid synthetic negatives and craft \emph{adversarial edits} to improve their ranks. The attacker does not add the decoy triple itself as an adversarial edit, rather chooses the adversarial edits that would improve the rank of a missing decoy triple through an inference pattern.

\paragraph{Threat Model:}
To ensure reliable vulnerability analysis, we use a white-box attack setting where the attacker has full knowledge of the target KGE model \citep{joseph_nelson_rubinstein_tygar_2019}. They cannot manipulate the model architecture or learned embeddings directly; but only through \emph{addition} of triples to the training data. We focus on adversarial \emph{additions} which are more challenging to design than adversarial deletions for sparse knowledge graphs\footnote{For every target triple, the possible number of adversarial additions in the neighbourhood of each entity are $\mathcal{E} \times \mathcal{R}$. For the benchmark dataset FB15k-237, this is of the order of millions; whereas the \emph{maximum} number of candidates for adversarial deletion are of the order of thousands.}.

As in prior studies \citep{pezeshkpour2019criage, zhang2019kgeattack}, the attacker is restricted to making edits only in the neighbourhood of target entities. They are also restricted to 1 decoy triple for each entity of the target triple.  
Furthermore, because of the use of filtered settings for KGE evaluation, the attacker cannot add the decoy triple itself to the training data (which intuitively would be a way to improve the decoy triple's rank).

% --------------------------------------------------------------------------------------------
% --------------------------------------------------------------------------------------------

\section{Poisoning Knowledge Graph Embeddings through Relation Inference Patterns }
Since the inference patterns on the knowledge graph specify a logic property between the relations, they can be expressed as Horn Clauses which is a subset of FOL formulae. For example, a property represented in the form  $\forall x,y : (x, \mathtt{owns}, y) \Rightarrow (y, \mathtt{is\_owned\_by}, x)$ means that two entities linked by relation $\mathtt{owns}$ are also likely to be linked by the inverse relation $\mathtt{is\_owned\_by}$. In this expression, the right hand side of the implication $\Rightarrow$ is referred to as the \emph{head} and the left hand side as the \emph{body} of the clause. 
Using such expressions, we define the three inference patterns used in our research.

\begin{definition}
    The \textbf{symmetry} pattern $\mathcal{P}_s$ is expressed as \(
        \forall x,y : (x, \mathtt{r}, y) \Rightarrow (y, \mathtt{r}, x)
    \). Here, the relation $\mathtt{r}$ is symmetric relation.
\end{definition}
\begin{definition}
    The \textbf{inversion} pattern $\mathcal{P}_i$ is expressed as 
    \(
        \forall x,y : (x, \mathtt{r_i}, y) \Rightarrow (y, \mathtt{r}, x)
    \). Here, the relations $\mathtt{r_i}$ and $\mathtt{r}$ are inverse of each other.
\end{definition}
\begin{definition}
    The \textbf{composition} pattern $\mathcal{P}_c$ is expressed as 
    \(
        \forall x,y,z : (x, \mathtt{r_1}, z) \wedge (z, \mathtt{r_2}, y) \Rightarrow (x, \mathtt{r}, y)
    \). Here, the relation $\mathtt{r}$ is a composition of $\mathtt{r_1}$ and $\mathtt{r_2}$ ; and the $\wedge$ is the conjunction operator from relational logic.
\end{definition}

The mapping \(\mathcal{G}: \mathcal{V \rightarrow \mathcal{E}} \) of variables $\mathcal{V}$ in the above expressions to entities $\mathcal{E}$ is called a grounding.
For example, we can map the logic expression $\forall x,y : (x, \mathtt{owns}, y) \Rightarrow (y, \mathtt{is\_owned\_by}, x)$ to the grounding $(Karl, \mathtt{owns}, Account42) \Rightarrow (Account42, \mathtt{is\_owned\_by}, Karl)$. Thus, a KGE model that captures the inversion pattern will assign a high prediction confidence to the head atom when the body of the clause exists in the graph.

In the above expressions, the decoy triple becomes the head atom and adversarial edits are the triples in the body of the expression.
Since the decoy triple is an object or subject side negative of the target triple, the attacker already knows the relation in the head atom. They now want to determine  (i) the adversarial relations in the body of the expression; (ii) the decoy entities which will most likely violate the inference pattern for the chosen relations and; (iii) the remaining entities in the body of the expression which will improve the prediction on the chosen decoy triple. Notice that the attacker needs all three steps for composition pattern only; for inversion pattern, only the first two steps are needed; and for symmetry pattern, only the second step is needed. 
Below we describe each step in detail. A computational complexity analysis of all the steps is available in Appendix \ref{apx:complexity_analysis}.

% -------------------------------------------------------------------------------------------------------------------

\subsection{Step1: Determine Adversarial Relations}
Expressing the relation patterns as logic expressions is based on relational logic and assumes that the relations are constants. Thus, we use an algebraic approach to determine the relations in the head and body of a clause.
Given the target relation $\mathtt{r}$, we determine the adversarial relations  using an algebraic model of inference \citep{yang2015distmult}.

\textbf{Inversion:} 
If an atom $(x, \mathtt{r}, y)$ holds true, then for the learned embeddings in multiplicative models, we can assume $\vec{e}_x \circ \vec{e}_{\mathtt{r}} \approx \vec{e}_y$; where $\circ$ denotes the Hadamard (element-wise) product. If the atom $(y, \mathtt{r_i}, x)$ holds true as well, then we can also assume $\vec{e}_y \circ \vec{e}_{\mathtt{r_i}} \approx \vec{e}_x$. Thus, \(\vec{e}_{\mathtt{r}} \circ \vec{e}_{\mathtt{r_i}} \approx \vec{1}\) for inverse relations $\mathtt{r}$ and $\mathtt{r_i}$ when embeddings are learned from multiplicative models. We obtain a similar expression  \(\vec{e}_{\mathtt{r}} + \vec{e}_{\mathtt{r_i}} \approx 0\) when embeddings are learned from additive models.

Thus, to determine adversarial relations for \emph{inversion} pattern, we use the pre-trained embeddings to select $\mathtt{r_i}$ that minimizes $\abs{\vec{e}_{\mathtt{r_i}} \vec{e}_{\mathtt{r}}^T - 1}$ for multiplicative models; and $\mathtt{r_i}$ that minimizes $\abs{\vec{e}_{\mathtt{r_i}} + \vec{e}_{\mathtt{r}}}$ for additive models.

\textbf{Composition}: If two atoms $(x, \mathtt{r_1}, y)$ and $(y, \mathtt{r_2}, z)$ hold true, then for multiplicative models, \(\vec{e}_x \circ \vec{e}_{\mathtt{r_1}} \approx \vec{e}_y\) and \(\vec{e}_y \circ \vec{e}_{\mathtt{r_2}} \approx \vec{e}_z \). Therefore, \(\vec{e}_x \circ (\vec{e}_{\mathtt{r_1}} \circ \vec{e}_{\mathtt{r_2}}) \approx \vec{e}_z\). Hence, relation $\mathtt{r}$ is a composition of $\mathtt{r_1}$ and $\mathtt{r_2}$ if $\vec{e}_{\mathtt{r_1}} \circ \vec{e}_{\mathtt{r_2}} \approx \vec{e}_{\mathtt{r}}$. Similarly, for embeddings from additive models, we can model composition as $\vec{e}_{\mathtt{r_1}} + \vec{e}_{\mathtt{r_2}} \approx \vec{e}_{\mathtt{r}}$.

Thus, to determine adversarial relations for \emph{composition} pattern, we use pre-trained embeddings to obtain all possible compositions of ($\mathtt{r_1}, \mathtt{r_2}$). For multiplicative models, we use $\vec{e}_{\mathtt{r_1}} \circ \vec{e}_{\mathtt{r_2}}$ and for additive models we use $\vec{e}_{\mathtt{r_1}} + \vec{e}_{\mathtt{r_2}}$. From these, we choose the relation pair for which the Euclidean distance between the composed relation embeddings and the target relation embedding $\vec{e}_{\mathtt{r}}$ is minimum.

% ------------------------------------------------------------------------------------------------------------------

\subsection{Step2: Determine Decoy Entities}
We consider three different heuristic approaches to select the decoy entity - soft truth score, ranks predicted by the KGE model and cosine distance. 

\paragraph{Soft Logical Modelling of Inference Patterns}
Once the adversarial relations are determined, we can express the grounding for symmetry, inversion and composition patterns for the decoy triples. We discuss only object side decoy triple for brevity - 
\begin{align*}
    \mathcal{G}_s: (o', \mathtt{r}, s) &\Rightarrow (s, \mathtt{r}, o') \\
    \mathcal{G}_i: (o', \mathtt{r_i},s) &\Rightarrow (s, \mathtt{r}, o') \\
    \mathcal{G}_c: (s, \mathtt{r_1}, o'') \wedge (o'', \mathtt{r_2}, o') &\Rightarrow (s, \mathtt{r}, o')
\end{align*}
If the model captures $\mathcal{P}_s$, $\mathcal{P}_i$ or $\mathcal{P}_c$ to assign high rank to the target triple, then the head atom $(s, \mathtt{r}, o')$ of a grounding that violates this pattern is a suitable decoy triple. Adding the body of this grounding to the knowledge graph would improve the model performance on decoy triple through $\mathcal{P}_s$, $\mathcal{P}_i$ or $\mathcal{P}_c$. 

To determine the decoy triple this way, we need a measure of the degree to which a grounding satisfies an inference pattern.
We call this measure the \emph{soft truth score} $\phi : \mathcal{G} \rightarrow [0,1]$ - it provides the truth value of a logic expression indicating the degree to which the expression is true. We model the soft truth score of grounded patterns using t-norm based fuzzy logics \citep{hajek1998tnormfuzzylogics}.

The score $f_{sro}$ of an individual atom (i.e. triple) is computed using the KGE model's scoring function. We use the sigmoid function $\sigma(x)=1/(1+\exp(-x))$ to map this score to a continuous truth value in the range $(0,1)$. Hence, the soft truth score for an individual atom is \(\phi(s,\mathtt{r},o) = \sigma(f_{sro})\).
The soft truth score for the grounding of a pattern can then be expressed through logical composition (e.g. $\wedge$ and $\Rightarrow$) of the scores of individual atoms in the grounding. We follow \citep{guo2016kale, guo2018ruge} and define the following compositions for logical conjunction ($\wedge$), disjunction ($\vee$), and negation ($\neg$):
\begin{align*}
  \phi(a \wedge b) & = \phi(a) \cdot \phi(b), \\
  \phi(a \vee b)   & = \phi(a) + \phi(b) - \phi(a) \cdot \phi(b), \\
  \phi(\neg a)     & = 1 - \phi(a).
\end{align*}

Here, $a$ and $b$ are two logical expressions, which can either be single triples or be constructed by combining triples with logical connectives. If $a$ is a single triple $(s,\mathtt{r},o)$, we have $\phi(a)=\phi(s,\mathtt{r},o)$. Given these compositions, the truth value of any logical expression can be calculated recursively \citep{guo2016kale, guo2018ruge}. 
 
Thus, we obtain the following soft truth scores for the groundings of symmetry, inversion and composition patterns $\mathcal{G}_s$, $\mathcal{G}_i$ and $\mathcal{G}_c$ - 
\begin{align*}
    \phi(\mathcal{G}_s) &= \phi(o',\mathtt{r},s) \cdot \phi(s,\mathtt{r},o') - \phi(o',\mathtt{r},s) + 1  \\
    \phi(\mathcal{G}_i) &= \phi(o',\mathtt{r_i},s) \cdot \phi(s,\mathtt{r},o') - \phi(o',\mathtt{r_i},s) + 1. \\
    \phi(\mathcal{G}_c) &= \phi(s,\mathtt{r_1},o'') \cdot \phi(o'',\mathtt{r_2},o') \cdot \phi(s,\mathtt{r},o') \\ &- \phi(s,\mathtt{r_1},o'') \cdot \phi(o'',\mathtt{r_2},o') + 1
\end{align*}

To select the decoy triple $(s,\mathtt{r},o')$ for symmetry and inversion, we score all possible groundings using $\phi(\mathcal{G}_s)$ and $\phi(\mathcal{G}_i)$. The head atom of grounding with minimum score is chosen as decoy triple.

For composition pattern, the soft truth score $\phi(\mathcal{G}_c)$ for candidate decoy triples $(s,\mathtt{r},o')$ contains two entities $(o',o'')$ to be identified. Thus, we use a greedy approach to select the decoy entity $o'$. We use the pre-trained embeddings to group the entities $o''$ into $k$ clusters using K-means clustering and determine a decoy entity with minimum soft truth score for each cluster. We then select the decoy entity $o'$ with minimum score across the $k$ clusters.

\paragraph{KGE Ranks:} 
We use the ranking protocol from KGE evaluation to rank the target triple against valid subject and object side negatives $(s',\mathtt{r},o)$ and $(s,\mathtt{r},o')$. For each side, we select the negative triple that is ranked just below the target triple (that is, $negative\_rank = target\_rank + 1$). These are suitable as decoy because their predicted scores are likely not very different from the target triple's score. Thus, the model's prediction confidence for these triples might be effectively manipulated through adversarial additions. This is in contrast to very low ranked triples as decoy; where the model has likely learnt a low score with high confidence.

\paragraph{Cosine Distance:}
A high rank for the target triple $(s,\mathtt{r},o)$ against queries $(s,\mathtt{r},?)$ and $(?,\mathtt{r},o)$ indicates that $\vec{e}_s, \vec{e}_o$ are similar to the embeddings of other subjects and objects related by $\mathtt{r}$ in the training data. Thus, a suitable heuristic for selecting decoy entities $s'$ and $o'$ is to choose ones whose embeddings are dissimilar to $\vec{e}_s, \vec{e}_o$. Since these entities are not likely to occur in the neighbourhood of $o$ and $s$, they will act adversarially to reduce the rank of target triple. 
Thus, we select decoy entities $s'$ and $o'$ that have maximum cosine distance from target entities $s$ and $o$ respectively.

% ---------------------------------------------------------------------------------------------------------------------

\subsection{Step3: Determine Adversarial Entities}
This step is only needed for the composition pattern because the body for this pattern has two adversarial triples.
Given the decoy triple in the head of the composition expression, we select the body of the expression that would maximize the rank of the decoy triple. We use the soft-logical model defined in Step 2 for selecting decoy triples. The soft truth score for composition grounding of decoy triple is given by $\phi(\mathcal{G}_t) = \phi(s,\mathtt{r_1},o'') \cdot \phi(o'',\mathtt{r_2},o') \cdot \phi(s,\mathtt{r},o') - \phi(s,\mathtt{r_1},o'') \cdot \phi(o'',\mathtt{r_2},o') + 1$. We select the entity $o''$ with maximum score because this entity satisfies the composition pattern for the decoy triple and is thus likely to improve the decoy triple's ranks on addition to the knowledge graph.

\begin{table}[]
    \centering
    \small
    \begin{tabular}{c c c c }
    \toprule
          \textbf{Adversarial Attack Step}    &      \textbf{Sym}   &   \textbf{Inv}   &   \textbf{Com}\\
    \midrule
        {Determine Adversarial Relations}  &   n/a       &   Alg   &   Alg  \\
    \midrule
         \multirow{3}{*}{\shortstack[l]{Determine Decoy Entities}}  &   Sft    &    Sft   &   Sft   \\
                                                &     Rnk   &    Rnk   &   Rnk  \\
                                                &     Cos   &   Cos   &   Cos  \\
    \midrule
        Determine Adversarial Entities                    &   n/a     &   n/a      &      Sft  \\
    \bottomrule
    \end{tabular}
    \caption{A summary of heuristic approaches used for different steps of the adversarial attack with symmetry (Sym), inversion (Inv) and composition (Com) pattern. Alg denotes the algebraic model for inference patterns; Sft denotes the soft truth score; Rnk denotes the KGE ranks; and Cos denotes the cosine distance.}
    \label{tab:attack_summary}
\end{table}

% -----------------------------------------------------------------------------------------------------------------------
% -----------------------------------------------------------------------------------------------------------------------

\section{Evaluation}
The aim of our evaluation is to assess the effectiveness of proposed attacks in \emph{degrading} the predictive performance of KGE models on missing triples that are predicted true.
We use the state-of-art evaluation protocol for data poisoning attacks \citep{xu2020advgraphsurvey}. We train a clean model on the original data; then generate the adversarial edits and add them to the dataset; and finally retrain a new model on this poisoned data. All hyperparameters for training on original and poisoned data remain the same. 

We evaluate four models with varying inductive abilities - DistMult, ComplEx, ConvE and TransE; on two publicly available benchmark datasets for link prediction\footnote{https://github.com/TimDettmers/ConvE}- WN18RR and FB15k-237. 
We filter out triples from the validation and test set that contain unseen entities.
To assess the attack effectiveness in \emph{degrading} performance on triples predicted as true, we need a set of triples that are predicted as true by the model.
Thus, we select as \emph{target triples}, a subset of the original test set where each triple is ranked $\leq$ 10 by the original model.
Table \ref{tab:data} provides an overview of dataset statistics and the number of target triples selected.

\begin{table}[]
\centering
\footnotesize
\setlength{\tabcolsep}{5pt}
\begin{tabular}{c  l ll}
    \toprule           
    \multicolumn{2}{l}{} & \textbf{WN18RR} &  \textbf{FB15k-237} \\ 
    \midrule
    \multicolumn{2}{l}{Entities}                  &  40,559   & 14,505 \\ 
    \multicolumn{2}{l}{Relations}                 &  11       & 237 \\ 
    \multicolumn{2}{l}{Training}                  & 86,835    &  272,115            \\ 
    \multicolumn{2}{l}{Validation}                &  2,824    &  17,526   \\ 
    \multicolumn{2}{l}{Test}                      &   2,924   &  20,438    \\ 
    \midrule
    \multirow{4}{*}{Target} 
    & DistMult  &   1,315   &  3,342    \\
    & ComplEx   & 1,369  &  3,930    \\
    & ConvE     &    1,247   &  4,711    \\
    & TransE    &    1,195   &   5,359   \\
    % && \\
    \bottomrule
\end{tabular}
\caption{\small Statistics for the datasets WN18RR and FB15k-237.
We removed triples from the validation and test set that contained unseen entities to ensure that we do not add new entities as adversarial edits. 
The numbers above (including the number of entities) reflect this filtering.}
\label{tab:data}
\end{table}

\paragraph{Baselines: }
We compare the proposed methods against the following baselines - 

\emph{Random\_n}: Random edits in the neighbourhood of each entity of the target triple.

\emph{Random\_g1}: Global random edits in the knowledge graph which are not restricted to the neighbourhood of entities in the target triple and have 1 edit per decoy triple (like symmetry and inversion).

\emph{Random\_g2}: Global random edits in the knowledge graph which are not restricted to the neighbourhood of entities in the target triple and have 2 edits per decoy triple (like composition).

\emph{Zhang et al.}: Poisoning attack from \cite{zhang2019kgeattack} for edits in the neighbourhood of subject of the target triple. We extend it for both subject and object to match our evaluation protocol. Further implementation details available in Appendix \ref{apx:ijcai_baseline}.

\emph{CRIAGE}: Poisoning attack from \citep{pezeshkpour2019criage}. We use the publicly available implementation and the default attack settings\footnote{https://github.com/pouyapez/criage}. The method was proposed for edits in the neighbourhood of object of the target triple. We extend it for both entities to match our evaluation protocol and to ensure fair evaluation.

\paragraph{Implementation:}
For every attack, we filter out adversarial edit candidates that already exist in the graph.
We also remove duplicate adversarial edits for different targets before adding them to the original dataset. 
For Step 2 of the composition attack with ground truth, we use the elbow method to determine the number of clusters for each model-data combination. Further details on KGE model training, computing resources and number of clusters are available in Appendix \ref{apx:implementation}. The source code to reproduce our experiments is available on GitHub\footnote{\url{https://github.com/PeruBhardwaj/InferenceAttack}}.

% -----------------------------------------------------------------------------------------------------------------------

\subsection{Results}
\label{sec:results}

\begin{table*}
\centering
\small
\setlength{\tabcolsep}{3.5pt}
\begin{tabular}{c  l  ll  ll   ll  ll }

\toprule
    
     & & \multicolumn{2}{c}{\textbf{DistMult}} & \multicolumn{2}{c}{\textbf{ComplEx}} & \multicolumn{2}{c}{\textbf{ConvE}} & \multicolumn{2}{c}{\textbf{TransE}} \\
   \cmidrule(lr){3-4}  \cmidrule(lr){5-6}  \cmidrule(lr){7-8} \cmidrule(lr){9-10} 
    & & \multicolumn{1}{l}{\textbf{MRR}}    & \multicolumn{1}{l}{\textbf{Hits@1}}  & \multicolumn{1}{l}{\textbf{MRR}}    & \multicolumn{1}{l}{\textbf{Hits@1}} & \multicolumn{1}{l}{\textbf{MRR}}    & \multicolumn{1}{l}{\textbf{Hits@1}} & \multicolumn{1}{l}{\textbf{MRR}}    & \multicolumn{1}{l}{\textbf{Hits@1}} \\
\midrule
    
     \textbf{Original}  &    & 0.90            &  0.85     &  0.89             &  0.84      &  0.92           &   0.89    & 0.36           &  0.03 \\
\midrule
    \multirow{5}{*}{\shortstack[l]{\textbf{Baseline} \\ \textbf{Attacks}}}
    & Random\_n     & 0.86 (-4\%)     &  0.83     &  0.84 (-6\%)      &    0.80    &  0.90 (-2\%)    &   0.88    &  0.28 (-20\%)  &   0.01 \\
    & Random\_g1    & 0.88            &  0.83     &  0.88             &   0.83     &  0.92           &   0.89    &  0.35          &   0.02 \\
    & Random\_g2    & 0.88            &  0.83     &  0.88             &   0.83     &  0.91           &    0.89   &  0.34          &   0.02 \\
\cline{2-10} \\[-7pt]
    & Zhang et al.     & 0.82 (-8\%)    &  0.81     &  0.76 (-14\%)    &    0.74    & 0.90 (-2\%)     &   0.87    &  \textbf{0.24 (-33\%)}     &   \textbf{0.01} \\
    & CRIAGE        & 0.87          &  0.84     &  -                &   -        & 0.90         &   0.88    &  -            &   - \\
\midrule
    \multirow{9}{*}{\shortstack[l]{\textbf{Proposed} \\ \textbf{Attacks}}}
    & Sym\_truth    & 0.66           &  0.40     &  \textbf{0.56 (-33\%)}  &  \textbf{0.24}      & \textbf{0.61 (-34\%)}   &    \textbf{0.28}    & 0.57      &  0.36 \\
    & Sym\_rank     & 0.61           &  0.32     &  \textbf{0.56 (-33\%)}  &  \textbf{0.24}      & 0.62                    &    0.31    & 0.25               &  0.02 \\
    & Sym\_cos      & \textbf{0.57 (-36\%)} &  \textbf{0.32}  &  0.62     &   0.43     &  0.67        &    0.44                        & \textbf{0.24 (-33\%)}  &  \textbf{0.01} \\
\cline{2-10} \\[-7pt] 
    & Inv\_truth   & 0.87           &   0.83     &  0.86                  &   0.80     & 0.90          &    0.87    &  0.34          &  0.03 \\
    & Inv\_rank    & 0.86           &   0.83     &  0.85                  &   0.80     & 0.89 (-4\%)   &    0.85    &  0.25          &  0.02 \\
    & Inv\_cos     & 0.83 (-8\%)     &  0.82     &  0.80 (-10\%)          &   0.79     & 0.90          &    0.88    &  0.25 (-30\%)  &  0.01 \\
\cline{2-10} \\[-7pt]
    
    & Com\_truth   & 0.86           &  0.83      &  0.86                &   0.81     &  0.89           &    0.86   &   0.53 (+49\%)    &  0.27 \\
    & Com\_rank    & 0.85 (-5\%)      &  0.80      &  0.83                &   0.77     &  0.89           &    0.84   &   0.57            &  0.32 \\
    & Com\_cos     & 0.86           &  0.77      &  0.82 (-8\%)         &   0.70     &  0.88(-4\%)     &    0.83   &   0.53 (+49\%)     &   0.27 \\
    
\bottomrule    

\end{tabular}
\caption{\small Reduction in MRR and Hits@1 due to different attacks on the \textbf{target split} of WN18RR. First block of rows are the baseline attacks with random edits; second block is state-of-art attacks; remaining are the proposed attacks. For each block, we report the \emph{best} relative percentage difference from original MRR; computed as $(original-poisoned)/original*100$. Lower values indicate better results; best results for each model are in bold. Statistics on the target split are in Table \ref{tab:data}.}
\label{tab:benchmark_mrr_WN18RR}
\end{table*}

Table \ref{tab:benchmark_mrr_WN18RR} and \ref{tab:benchmark_mrr_FB15k-237} show the reduction in MRR and Hits@1 due to different attacks on the WN18RR and FB15k-237 datasets. 
We observe that the proposed adversarial attacks outperform the random baselines and the state-of-art poisoning attacks for all KGE models on both datasets. 

We see that the attacks based on symmetry inference pattern perform the best across all model-dataset combinations. This indicates the sensitivity of KGE models to symmetry pattern. For DistMult, ComplEx and ConvE, this sensitivity can be explained by the symmetric nature of the scoring functions of these models. That is, the models assign either equal or similar scores to triples that are symmetric opposite of each other. In the case of TransE, the model's sensitivity to symmetry pattern is explained by the translation operation in scoring function. The score of target $(s,\mathtt{r},o)$ is a translation from subject to object embedding through the relation embedding. Symmetry attack adds the adversarial triple $(o',\mathtt{r},s)$ where the relation is same as the target relation, but target subject is the object of adversarial triple. Now, the model learns the embedding of $s$ as a translation from $o'$ through relation $\mathtt{r}$. This adversarially modifies the embedding of $s$ and in turn, the score of $(s,\mathtt{r},o)$. 

We see that inversion and composition attacks also perform better than baselines in most cases, but not as good as symmetry. This is particularly true for FB15k-237 where the performance for these patterns is similar to random baselines. For the composition pattern, it is likely that the model has stronger bias for shorter and simpler patterns like symmetry and inversion than for composition. This makes it harder to deceive the model through composition than through symmetry or inverse. Furthermore, FB15k-237 has high connectivity \citep{dettmers2018conve} which means that a KGE model relies on a high number of triples to learn target triples' ranks. Thus, poisoning KGE models for FB15k-237 will likely require more adversarial triples per target triple than that considered in this research. 

The inversion pattern is likely ineffective on the benchmark datasets because these datasets do not have any inverse relations \citep{dettmers2018conve,toutanova2015observed}. This implies that our attacks cannot identify the inverse of the target triple's relation in Step 1. We investigate this hypothesis further in Appendix \ref{apx:WN18_analysis}, and evaluate the attacks on WN18 dataset where the inverse relations have not been filtered out. This means that the KGE model can learn the inversion pattern and the inversion attacks can identify the inverse of the target relation. In this setting, we find that the inversion attacks outperform other attacks against ComplEx on WN18, indicating the sensitivity of ComplEx to the inversion pattern when the dataset contains inverse relations.

An exception in the results is the composition pattern on TransE where the model performance improves instead of degrading on the target triples. This is likely due to the model's sensitivity to composition pattern such that adding this pattern improves the performance on all triples, including target triples. To verify this, we checked the change in ranks of decoy triples and found that composition attacks on TransE improve these ranks too. Results for this experiment are available in Appendix \ref{apx:decoy_analysis}. This behaviour of composition also indicates that the selection of adversarial entities in Step 3 of the composition attacks can be improved. It also explains why the increase is more significant for WN18RR than FB15k-237 - WN18RR does not have any composition relations but FB15k-237 does; so adding these to WN18RR shows significant improvement in performance. 
We aim to investigate these and more hypotheses about the proposed attacks in future work.

\begin{table*}[]
\centering
\small
\setlength{\tabcolsep}{3.5pt}
\begin{tabular}{c  l  ll  ll   ll  ll }
\toprule
    & & \multicolumn{2}{c}{\textbf{DistMult}} & \multicolumn{2}{c}{\textbf{ComplEx}} & \multicolumn{2}{c}{\textbf{ConvE}} & \multicolumn{2}{c}{\textbf{TransE}} \\
   \cmidrule(lr){3-4}  \cmidrule(lr){5-6}  \cmidrule(lr){7-8} \cmidrule(lr){9-10} 
    & & \textbf{MRR}   & \textbf{Hits@1}  & \textbf{MRR}   & \textbf{Hits@1} & \textbf{MRR}   & \textbf{Hits@1} & \textbf{MRR}   & \textbf{Hits@1} \\
\midrule
     \textbf{Original}  & & 0.61              &  0.38     & 0.61       &   0.45             & 0.61              &   0.45      &    0.63               &  0.48   \\
\midrule
    \multirow{5}{*}{\shortstack[l]{\textbf{Baseline} \\ \textbf{Attacks}}}
    & Random\_n  & 0.54 (-11\%)    &  0.40     & 0.54  (-12\%)  &    0.40          & 0.56 (-8\%)       &   0.41       &   0.60 (-4\%)       & 0.45   \\
    & Random\_g1 & 0.54            &  0.40     & 0.55           &   0.41           & 0.57              &   0.43       &    0.62             &  0.46   \\
    & Random\_g2  & 0.55           &  0.41     & 0.55           &   0.40           & 0.57              &   0.42        &  0.61             &  0.46   \\
\cline{2-10} \\[-7pt]
    & Zhang et al.   & 0.53 (-13\%)     &  0.39     & 0.51 (-16\%)    &  0.38            & 0.54  (-11\%)            &  0.39       &  0.57 (-10\%)       &  0.42  \\
    & CRIAGE     & 0.54     &  0.41     &  -              &  -               & 0.56        &  0.41       & -                     &  -   \\
\midrule
    \multirow{9}{*}{\shortstack[l]{\textbf{Proposed} \\ \textbf{Attacks}}}
    & Sym\_truth &  0.51            & 0.36       &  0.56         &   0.41             & \textbf{0.51} (\textbf{-17\%})       &   \textbf{0.34}      & 0.62          &  0.48  \\
    & Sym\_rank  &  0.53            &  0.39      &  0.53         & \textbf{0.38}      &  0.55          &  0.38        & \textbf{0.53} (\textbf{-16\%})       &   \textbf{0.36} \\
    & Sym\_cos   &  \textbf{0.46} (\textbf{-25\%}) &  \textbf{0.31} & \textbf{0.51} (\textbf{-17\%})  &  \textbf{0.38}     
               & 0.52            &   0.37        & 0.55         &  0.40  \\
\cline{2-10} \\[-7pt]  
    & Inv\_truth  & 0.55            & 0.41       &   0.54          &   0.40        &  0.56              &  0.41        & 0.62               & 0.46   \\
    & Inv\_rank   & 0.56            &  0.43      &   0.55          &   0.40        &  0.55  (-9\%)      &   0.40       & 0.58 (-8\%)       &  0.42  \\
    & Inv\_cos    & 0.54 (-11\%)      &  0.40     &  0.53 (-14\%)  &   0.39        &  0.56              &   0.42        & 0.59              &  0.44   \\
\cline{2-10} \\[-7pt]
    
    & Com\_truth  & 0.56          &  0.42      &  0.55             &   0.41       &   0.57            &    0.43       & 0.65                &  0.51  \\
    & Com\_rank   & 0.56 (-8\%)      & 0.42      & 0.55 (-11\%)    &   0.40       &   0.56 (-8\%)     &   0.41        & 0.69                &  0.48   \\
    & Com\_cos    & 0.56 (-8\%)      &  0.43     & 0.56            &   0.42      &    0.56            &   0.42        & 0.63 (0\%)       &  0.49   \\
    
\bottomrule    

\end{tabular}
\caption{\small Reduction in MRR and Hits@1 due to different attacks on the \textbf{target split} of FB15k-237. For each block of rows, we report the \emph{best} relative percentage difference from original MRR; computed as $(original-poisoned)/original*100$. Lower values indicate better results; best results for each model are in bold. Statistics on the target split are in Table \ref{tab:data}.}
\label{tab:benchmark_mrr_FB15k-237}
\end{table*}

% ----------------------------------------------------------------------------------------------------------------------------------------------
% ----------------------------------------------------------------------------------------------------------------------------------------------

\section{Related Work}

KGE models can be categorized into tensor factorization models like DistMult \citep{yang2015distmult} and ComplEx \citep{trouillon2016complex}, neural architectures like ConvE \citep{dettmers2018conve} and translational models like TransE \citep{bordes2013transe}. We refer the reader to \citep{cai2018comprehensive} for a comprehensive survey.
Due to the black-box nature of KGE models, there is an emerging literature on understanding these models. 
\citep{pezeshkpour2019criage} and \citep{zhang2019kgeattack} are most closely related to our work as they propose other data poisoning attacks for KGE models. 

\citet{minervini2017adversarialsets} and \citet{cai2018kbgan} use adversarial regularization in latent space and adversarial training to improve predictive performance on link prediction. But these adversarial samples are not in the input domain and aim to improve instead of degrade model performance.
Poisoning attacks have also been proposed for models for undirected and single relational graph data like Graph Neural Networks ~\citep{zugner2018nettack, dai2018adversarialgcn} and Network Embedding models \citep{bojchevski2019adversarialnetworkembedding}. A survey of poisoning attacks for graph data is available in \citep{xu2020advgraphsurvey}. But the attacks for these models cannot be applied directly to KGE models because they require gradients of a dense adjacency matrix.

In the literature besides adversarial attacks, \citet{lawrence2020gradientrollback}, \citet{nandwani2020oxkbc} and \citet{zhang2019interaction} generate post-hoc explanations to understand KGE model predictions. \citet{trouillon2019inductive} study the inductive abilities of KGE models as binary relation properties for controlled inference tasks with synthetic datasets. \citet{allen2021interpreting} interpret the structure of knowledge graph embeddings by comparison with word embeddings. On the theoretical side, \citet{wang2018multi} study the expressiveness of various bilinear KGE models and \citet{basulto2018ontologyembedding} study the ability of KGE models to learn hard rules expressed as ontological knowledge.

The soft-logical model of inference patterns in this work is inspired by the literature on injecting logical rules into KGE models. \citet{guo2016kale} and \citet{guo2018ruge} enforce soft logical rules by modelling the triples and rules in a unified framework and jointly learning embeddings from them. 
Additionally, our algebraic model of inference patterns, which is used to select adversarial relations, is related to approaches for graph traversal in latent vector space discussed in \citet{yang2015distmult, guu2015traversing, arakelyan2021complexqueryanswering}.

% ---------------------------------------------------------------------------------------------------------------------------------
% ---------------------------------------------------------------------------------------------------------------------------------

\section{Conclusion}
We propose data poisoning attacks against KGE models based on inference patterns like symmetry, inversion and composition. Our experiments show that the proposed attacks outperform the state-of-art attacks. 
Since the attacks rely on relation inference patterns, they can also be used to understand the KGE models. This is because if a KGE model is sensitive to a relation inference pattern, then that pattern should be an effective adversarial attack. 
We observe that the attacks based on symmetry pattern generalize across all KGE models which indicates their sensitivity to this pattern. 

In the future, we aim to investigate hypotheses about the effect of input graph connectivity and existence of specific inference patterns in datasets. 
We note that such investigation of inference pattern attacks will likely be influenced by the choice of datasets. In this paper, we have used benchmark datasets for link prediction. While there are intuitive assumptions about the inference patterns on these datasets, there is no study that formally measures and characterizes the existence of these patterns. This makes it challenging to verify the claims made about the inductive abilities of KGE models, not only by our proposed attacks but also by new KGE models proposed in the literature. 

Thus, a promising step in understanding knowledge graph embeddings is to propose datasets and evaluation tasks that test varying degrees of specific inductive abilities. 
These will help evaluate new models and serve as a testbed for poisoning attacks. Furthermore, specifications of model performance on datasets with different inference patterns will improve the usability of KGE models in high-stake domains like healthcare and finance.

In addition to understanding model behaviour, the sensitivity of state-of-art KGE models to simple inference patterns indicates that these models can introduce security vulnerabilities in pipelines that use knowledge graph embeddings. Thus, another promising direction for future work is towards mitigating the security vulnerabilities of KGE models. Some preliminary ideas for this research can look into adversarial training; or training an ensemble of different KGE scoring functions; or training an ensemble from subsets of the training dataset. Since our experiments show that state-of-art KGE models are sensitive to symmetry pattern, we call for future research to investigate neural architectures that generalize beyond symmetry even though their predictive performance for link prediction on benchmark datasets might not be the best.

% --------------------------------------------------------------------------------------------------
% --------------------------------------------------------------------------------------------------

\section*{Acknowledgements}
This research was conducted with the financial support of Accenture Labs and Science Foundation Ireland (SFI) at the ADAPT SFI Research Centre at Trinity College Dublin.  The ADAPT SFI Centre for Digital Content Technology is funded by Science Foundation Ireland through the SFI Research Centres Programme and is co-funded under the European Regional Development Fund (ERDF) through Grant No. 13/RC/2106\_P2.

% ---------------------------------------------------------------------------------------------------

\section*{Broader Impact}
We study the problem of generating data poisoning attacks on KGE models. Data poisoning attacks identify the vulnerabilities in learning algorithms that could be exploited by an adversary to manipulate the model's behaviour \citep{joseph_nelson_rubinstein_tygar_2019, biggio2018wild}. Such manipulation can lead to unintended model behaviour and failure. Identifying these vulnerabilities for KGE models is critical because of their increasing use in domains that need high stakes decision making like heathcare \citep{bendsten2019astrazeneca} and finance \citep{hogan2020knowledgegraphs, noy2019knowledgegraphs}. 
In this way, our research is directed towards minimizing the negative consequences of deploying state-of-art KGE models in our society. This honours the ACM code of Ethics of contributing to societal well-being and acknowledging that all people are stakeholders in computing. At the same time, we aim to safeguard the KGE models against potential harm from adversaries and thus honour the ACM code of avoiding harm due to computing systems.

Arguably, because we study vulnerabilities by attacking the KGE models, the proposed attacks can be used by an actual adversary to manipulate the model behaviour of deployed systems. This paradox of an arms race is universal across security research \citep{biggio2018wild}. For our research, we have followed the principle of proactive security as recommended by \citet{joseph_nelson_rubinstein_tygar_2019} and \citet{biggio2018wild}. As opposed to reactive security measures where learning system designers develop countermeasures after the system is attacked, a proactive approach \emph{anticipates} such attacks, simulates them and designs countermeasures before the systems are deployed. Thus, by revealing the vulnerabilities of KGE models, our research provides an opportunity to fix them.

Besides the use case of security, our research can be used in understanding the inductive abilities of KGE models, which are black-box and hard to interpret. We design attacks that rely on the inductive assumptions of a model to be able to deceive that model. Thus, theoretically, the effectiveness of attacks based on one inference pattern over another indicates the model's reliance on one inference pattern over another. However, as we discussed in our paper, realistically, it is challenging to make such claims about the inductive abilities of KGE models because the inference patterns in benchmark datasets are not well defined. 

Thus, we would encourage further work to evaluate our proposed attacks by designing benchmark tasks and datasets that measure specific inductive abilities of models. This will not only be useful for evaluating the proposed attacks here, but also for understanding the inductive abilities of existing KGE models. This in turn, can guide the community to design better models. In this direction, we encourage researchers proposing new KGE models to evaluate not only the predictive performance on benchmark datasets, but also the claims made on inductive abilities of these models and their robustness to violations of these implicit assumptions.

% -------------------------------------------------------------------------------

\bibliographystyle{acl_natbib}
\bibliography{acl2021}

% --------------------------------------------------------------------------------
% --------------------------------------------------------------------------------

\appendix
\section*{Appendix}

\section{Computational Complexity Analysis}
\label{apx:complexity_analysis}
Lets say \( \mathcal{E} \) is the set of entities and \( \mathcal{R} \) is the set of relations. 
The number of target triples to attack is $t$ and the specific target triple is $(s,\mathtt{r},o)$.
Here, we discuss the computational complexity of the three steps of the proposed attacks -
\paragraph{Determine Adversarial Relations:} In this step, we determine the inverse relation or the composition relation of a target triple. To select inverse relation, we need $\mathcal{R}$ computations for every target triple. Selecting composition relation requires the composition operation $\mathcal{R}^2$ times per target triple. To avoid repetition, we pre-compute the inverse and composition relations for all target triples. This gives the complexity $\mathcal{O}(\mathcal{R}^2)$ for inverse relation. For composition relation, we compute compositions of all relation pairs and then select the adversarial pair by comparison with target relation. This gives $\mathcal{O}(\mathcal{R}^2 + \mathcal{R})$ complexity for composition. 

\paragraph{Determine Decoy Entity:} The three heuristics to compute the decoy entity are soft-truth score, KGE ranks and cosine distance. For symmetry and inversion, the soft truth score requires 2 forward calls to the model for one decoy entity. For composition, if the number of clusters is $k$, the soft truth score requires $3k$ forward calls to the model. To select decoy entities based on KGE ranks, we require one forward call for each decoy entity. For cosine distance, we compute the similarity of $s$ and $o$ to all entities via two calls to Pytorch's $\mathtt{F.cosine\_similarity}$. Once the heuristic scores are computed, there is an additional complexity of $\mathcal{O}(\mathcal{E})$ to select the entity with minimum score. Thus, the complexity for decoy selection is $\mathcal{O}(t\mathcal{E})$ for all heuristics except soft truth score on composition where it is $\mathcal{O}(kt\mathcal{E})$.

\paragraph{Determine Adversarial Entity:} This step requires three forward calls to the KGE model because the ground truth score needs to be computed. Thus, the complexity for this step is $\mathcal{O}(t\mathcal{E})$.

Based on the discussion above, the overall computational complexity is $\mathcal{O}(t\mathcal{E})$ for symmetry attacks and $\mathcal{O}(\mathcal{R}^2 + t\mathcal{E})$ for inversion attacks. For composition attacks, it is $\mathcal{O}(\mathcal{R}^2 + \mathcal{R} + kt\mathcal{E})$ for soft truth score and $\mathcal{O}(\mathcal{R}^2 + \mathcal{R} + t\mathcal{E})$ for KGE ranks and cosine distance.

% -------------------------------------------------------------------------------------------------------
% -------------------------------------------------------------------------------------------------------

\section{Implementation Details}
\label{apx:implementation}
\subsection{Training KGE models} 
Our codebase\footnote{\url{https://github.com/PeruBhardwaj/InferenceAttack}} for KGE model training is based on the codebase from \citep{dettmers2018conve}\footnote{https://github.com/TimDettmers/ConvE}. We use the 1-K training protocol but without reciprocal relations. Each training step alternates through batches of (s,r) and (o,r) pairs and their labels. The model implementation uses an if-statement for the forward pass conditioned on the input batch mode. 

For TransE scoring function, we use L2 norm and a margin value of 9.0. The loss function used for all models is Pytorch's BCELosswithLogits. For regularization, we use label smoothing and L2 regularization for TransE; and input dropout with label smoothing for remaining models. We also use hidden dropout and feature dropout for ConvE.  

We do not use early stopping to ensure same hyperparameters for original and poisoned KGE models. We used an embedding size of 200 for all models on both datasets. For ComplEx, this becomes an embedding size of 400 because of the real and imaginary parts of the embeddings. All hyperparameters are tuned manually based on suggestions from state-of-art implementations of KGE models \citep{ruffinelli2020olddognewtricks, dettmers2018conve}. 
The hyperparameter values for all model dataset combinations are available in the codebase. 
Table \ref{tab:original_mrr} shows the MRR and Hits@1 for the original KGE models on WN18RR and FB15k-237. 

For re-training the model on poisoned dataset, we use the same hyperparameters as the original model.
We run all model training, adversarial attacks and evaluation on a shared HPC cluster with Nvidia RTX 2080ti, Tesla K40 and V100 GPUs.

\begin{table}[]
    \centering
    \small
    \begin{tabular}{c  c c   c c}
    \toprule
                  &  \multicolumn{2}{c}{\textbf{WN18RR}}       &   \multicolumn{2}{c}{\textbf{FB15k-237}}   \\
                  \cmidrule(lr){2-3}  \cmidrule(lr){4-5}  
                  &  \textbf{MRR}   &  \textbf{Hits@1}    &   \textbf{MRR}   &  \textbf{Hits@1}  \\
    \midrule
         DistMult &     0.42   &   0.39    &  0.27  &  0.19  \\
         ComplEx  &     0.43   &   0.40    &  0.24  &  0.20   \\
         ConvE    &     0.43   &   0.40    &  0.32  &  0.23  \\
         TransE   &     0.19   &   0.02    &  0.34  &  0.25  \\
    \bottomrule
    \end{tabular}
    \caption{MRR and Hits@1 results for original KGE models on WN18RR and FB15k-237}
    \label{tab:original_mrr}
\end{table}

\begin{figure*}[t]
    \centering
    \begin{subfigure}[htb]{0.30\textwidth}
        \includegraphics[width=1\textwidth]{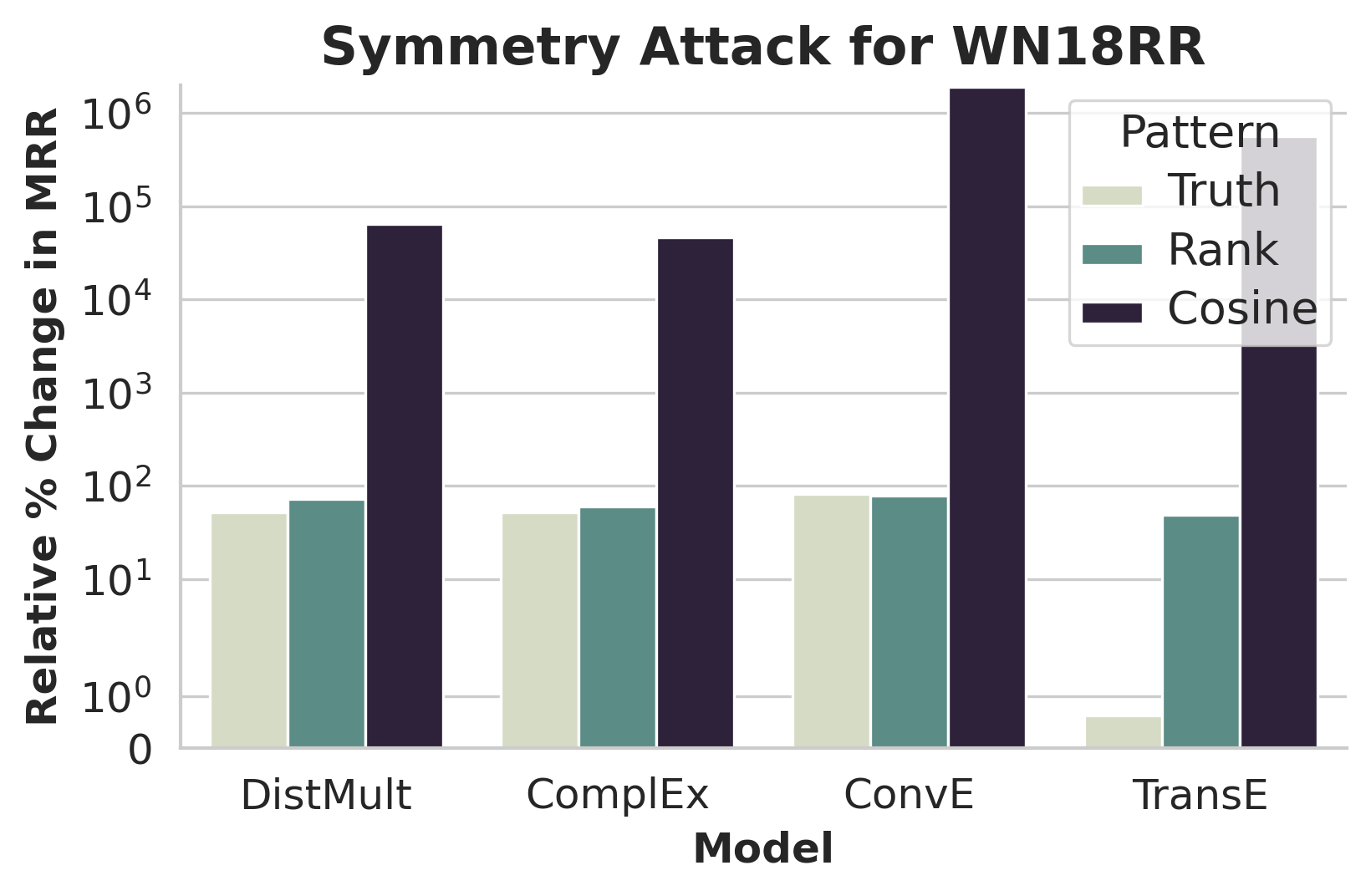}
    \end{subfigure}
    \quad
    \begin{subfigure}[htb]{0.30\textwidth}
        \includegraphics[scale=0.5,width=1\textwidth]{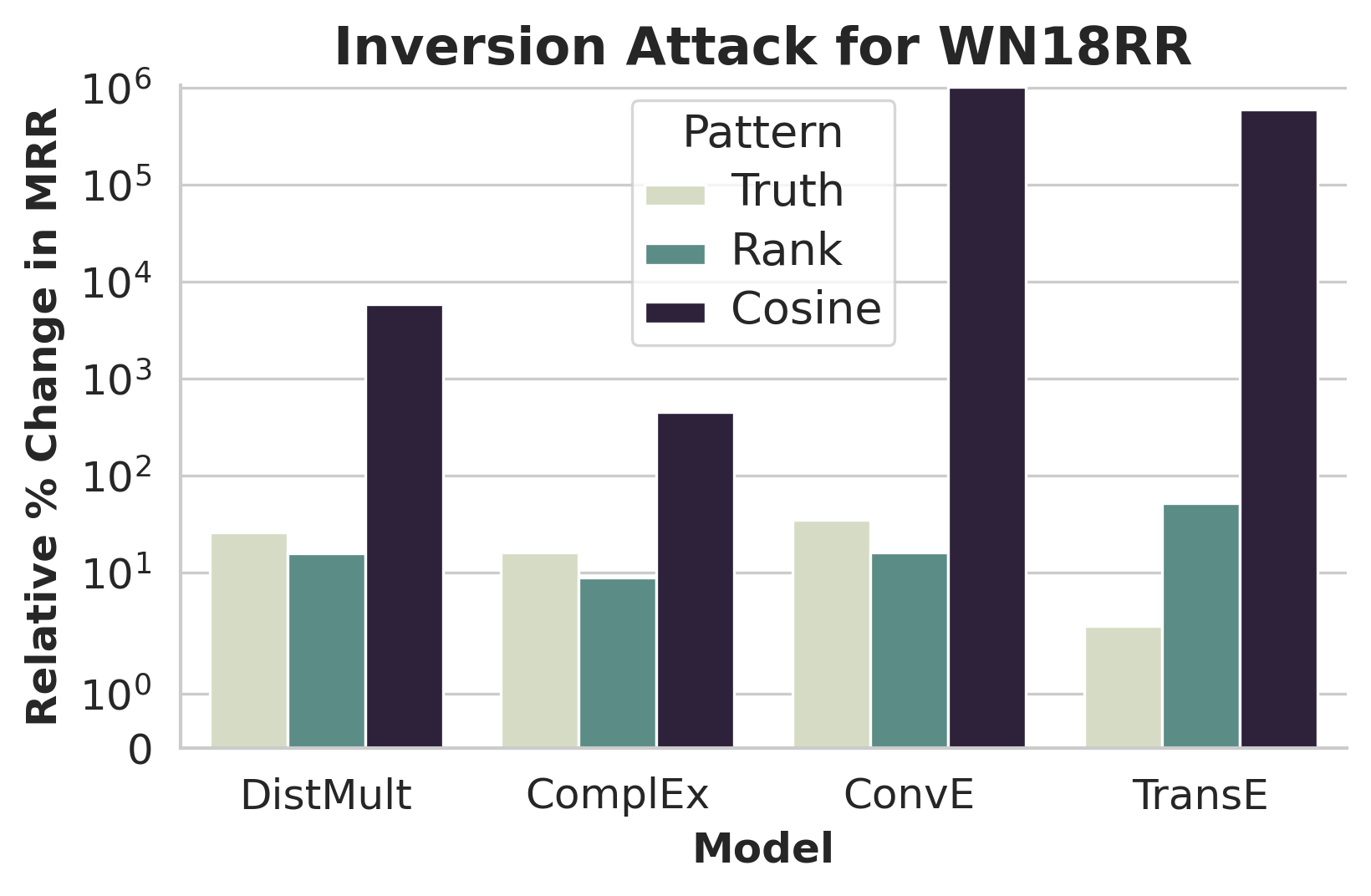}
    \end{subfigure}
    \quad
    \begin{subfigure}[htb]{0.30\textwidth}
        \includegraphics[width=1\textwidth]{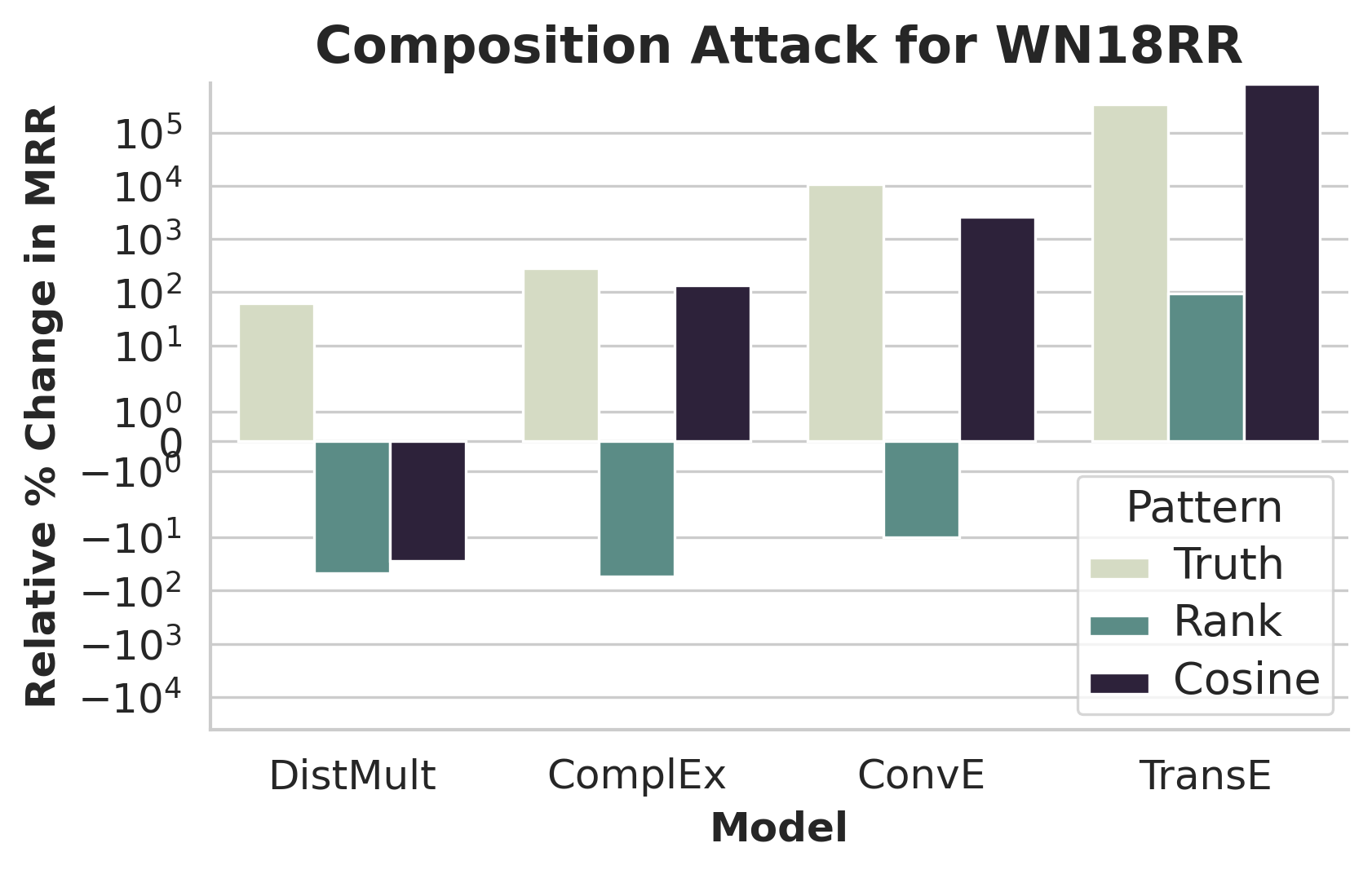}
    \end{subfigure} 
    \newline
    \newline
    \newline
    \begin{subfigure}[htb]{0.30\textwidth}
        \includegraphics[width=1\textwidth]{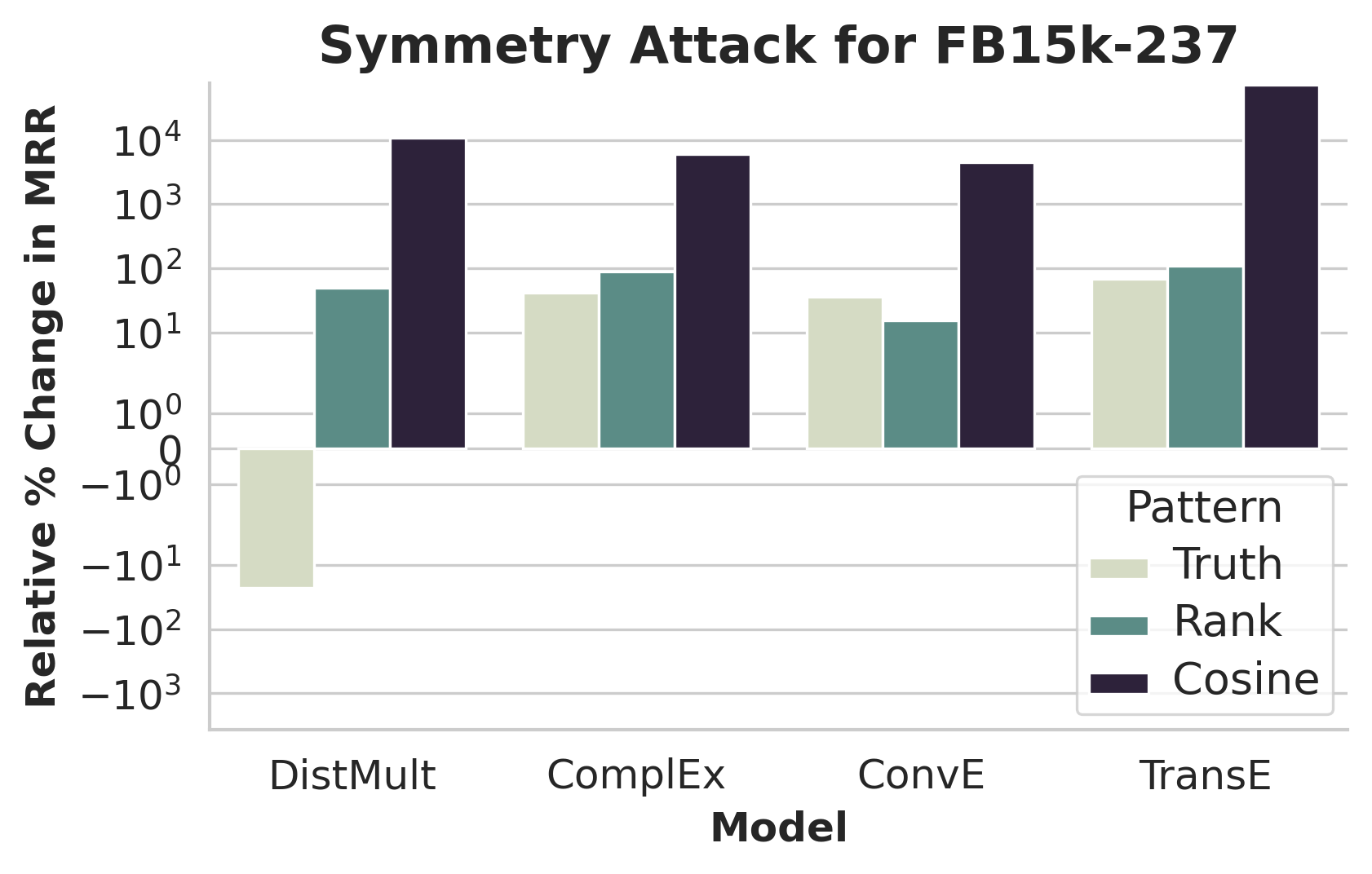}
    \end{subfigure}
    \quad
    \begin{subfigure}[htb]{0.30\textwidth}
        \includegraphics[width=1\textwidth]{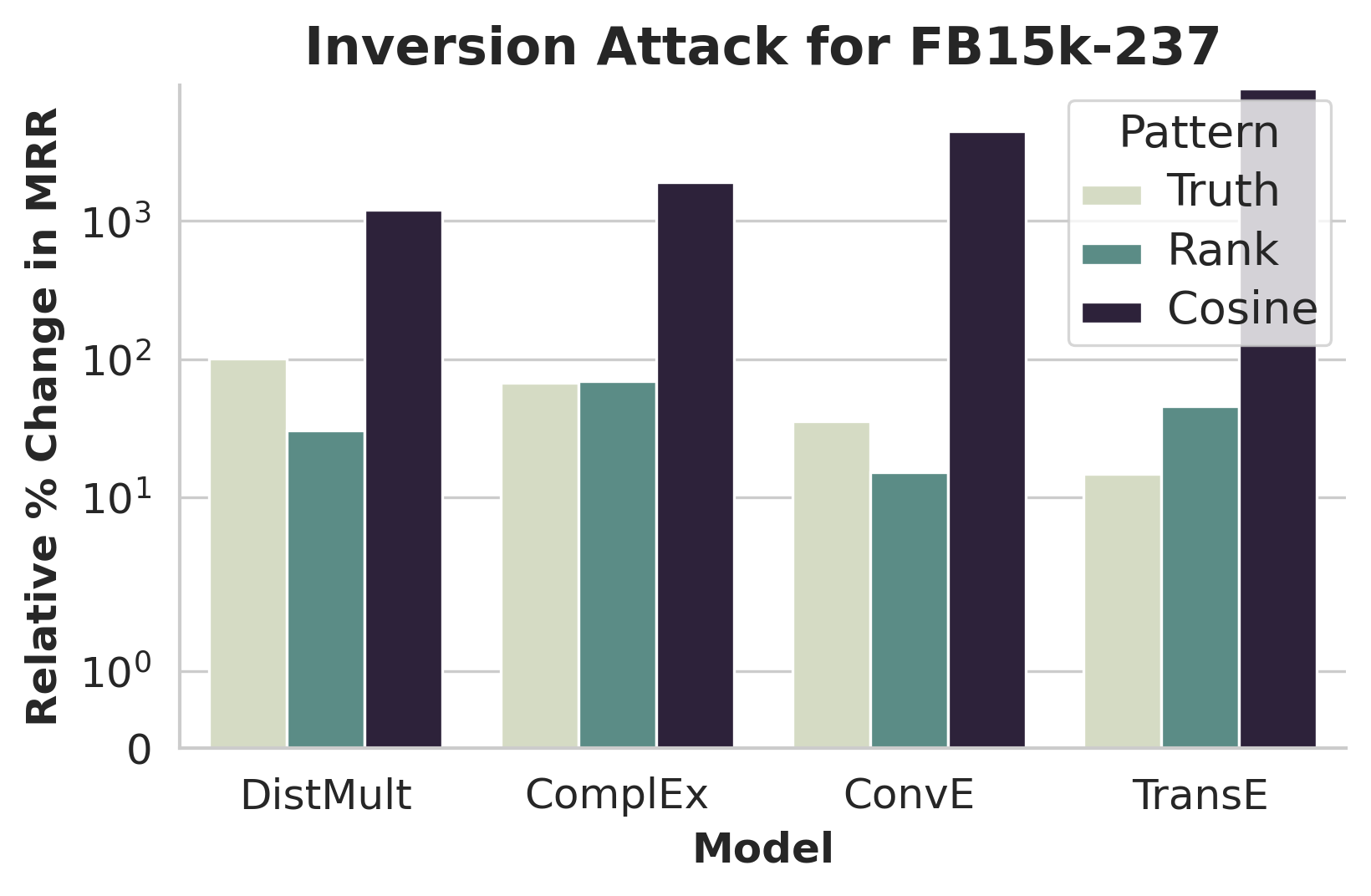}
    \end{subfigure}
    \quad
    \begin{subfigure}[htb]{0.30\textwidth}
        \includegraphics[width=1\textwidth]{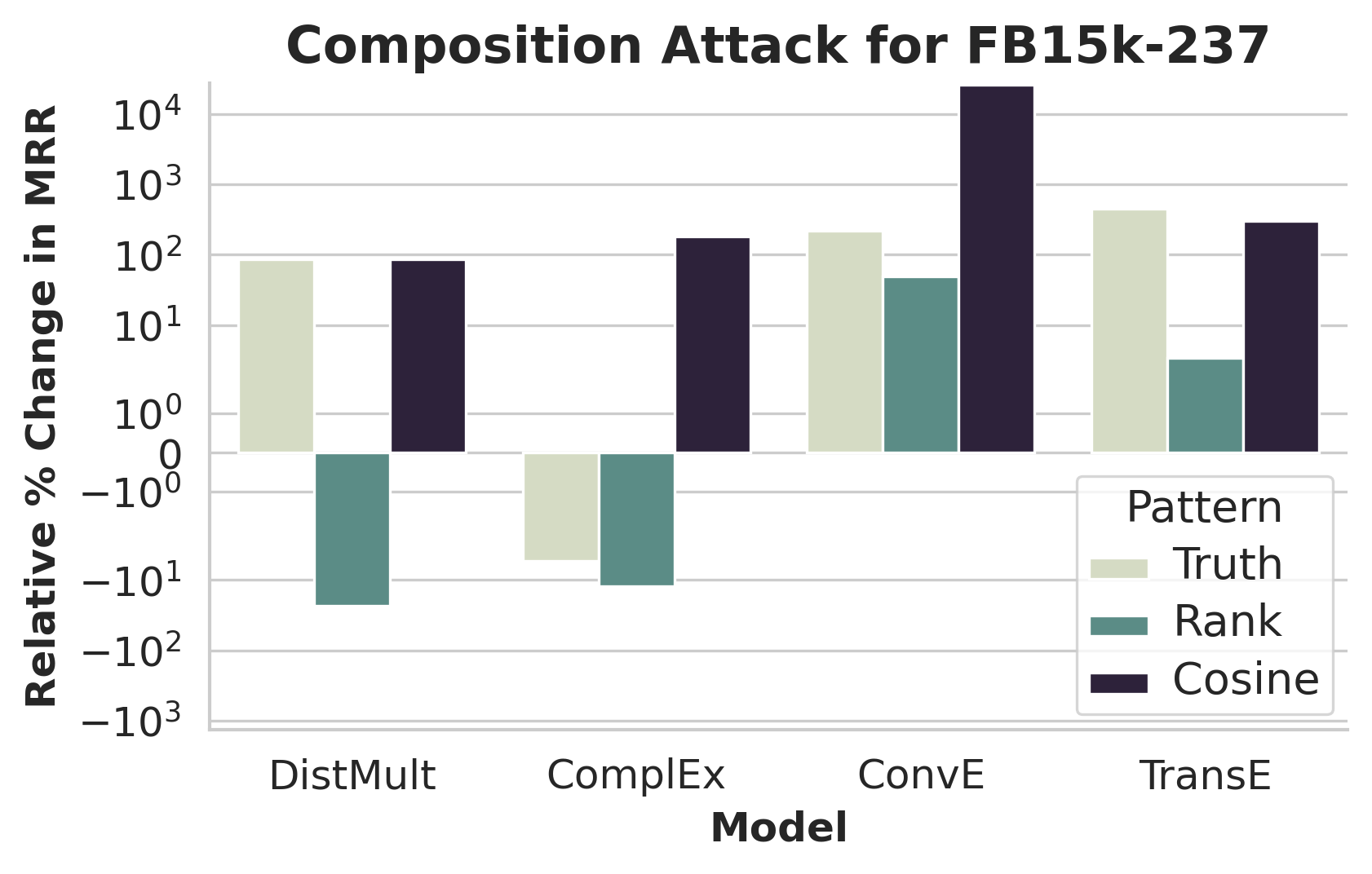}
    \end{subfigure}
    \caption{Mean of the relative increase in MRR of object and subject side \textbf{decoy triples} due to proposed attacks on WN18RR and FB15k-237. The increase is computed relative to original MRR of decoy triples as $(poisoned - original)/original$. The scale on y-axis is symmetric log scale. Higher values are better; as they show the effectiveness of attack in improving decoy triples' ranks relative to their original ranks.}\label{fig:decoy_graphs}
\end{figure*}

% ---------------------------------------------------------------------------------------------------------------

\subsection{Baseline Implementation Details}
\label{apx:ijcai_baseline}
One of the baselines in our evaluation is the attack from \cite{zhang2019kgeattack}. It proposed edits in the neighbourhood of subject of the target triple. We extend it for both subject and object to match our evaluation protocol. Since no public implementation is available, we implement our own.

The attack is based on computing a perturbation score for all possible candidate additions. Since the search space for candidate additions is of the order $\mathcal{E} \times \mathcal{R}$, the attack uses random down sampling to filter out the candidates. The percent of triples down sampled are not reported in the original paper and the implementation is not available. So, in this paper, we pick a high and a low value of the percentage of triples down sampled and generate adversarial edits for both fractions. 
The high and low percent values that were used to select candidate adversarial additions for WN18RR are DistMult: (20.0, 5.0); ComplEx: (20.0, 5.0); ConvE: (2.0, 0.1); TransE: (20.0, 5.0). For FB15k-237, these values are DistMult: (20.0, 5.0); ComplEx: (15.0, 5.0); ConvE: (0.3, 0.1); TransE: (20.0, 5.0)

\begin{table}[]
    \centering
    \small
    \begin{tabular}{c  ccc   }
    \toprule
                  &  \multicolumn{3}{c}{\textbf{WN18RR}}         \\
                  \cmidrule(lr){2-4} 
                  &  \textbf{Original}   &  \textbf{High}    &   \textbf{Low}     \\
    \midrule
         DistMult &     0.90      &   0.82        &   0.83        \\
         ComplEx  &     0.89      &   0.76        &   0.79       \\
         ConvE    &     0.92      &   0.90       &   0.90       \\
         TransE   &     0.36      &   0.25        &   0.24        \\
         
    \midrule
                 &   \multicolumn{3}{c}{\textbf{FB15k-237}}   \\
                  \cmidrule(lr){2-4}
                  &  \textbf{Original}   &  \textbf{High}    &   \textbf{Low} \\
    \midrule
        DistMult    &  0.61     &  0.55    &   0.53     \\
        ComplEx     &  0.61     &  0.51    &   0.52     \\
        ConvE       &  0.61     &  0.54     &   0.54    \\
        TransE      &  0.63     &  0.57    &   0.57     \\
                  
    \bottomrule
    \end{tabular}
    \caption{MRR of KGE models trained on original datasets and poisoned datasets from the attack in \citet{zhang2019kgeattack}. High, Low indicate the high and low percentage of candidates used for attack.}
    \label{tab:ijcai_results}
\end{table}

Thus, we generate two poisoned datasets from the attack - one that used a high number of candidates and another that used a low number of candidates. We train two separate KGE models on these datasets to assess attack performance. Table \ref{tab:ijcai_results} shows the MRR of the original model; and poisoned KGE models from attack with high and low downsampling percents. The results reported for this attack's performance in Section \ref{sec:results} are the better of the two results (which show more degradation in performance) for each combination.

% --------------------------------------------------------------------------------------------------------------

\subsection{Attack Implementation Details}

Our proposed attacks involve three steps to generate the adversarial additions for all target triples. For step1 of selection of adversarial relations, we pre-compute the inversion and composition relations for all target triples. Step2 and Step3 are computed for each target triple in a \emph{for} loop. These steps involve forward calls to KGE models to score adversarial candidates. For this, we use a vectorized implementation similar to KGE evaluation protocol. We also filter out the adversarial candidates that already exist in the training set. We further filter out any duplicates from the set of adversarial triples generated for all target triples. 

For the composition attacks with soft-truth score, we use the KMeans clustering implementation from \(\mathtt{scikit-learn}\). We use the elbow method on the grid [5, 20, 50, 100, 150, 200, 250, 300, 350, 400, 450, 500] to select the number of clusters. The number of clusters selected for WN18RR are DistMult: 300, ComplEx: 100, ConvE: 300, TransE: 50. For FB15k-237, the numbers are DistMult: 200, ComplEx: 300, ConvE: 300, TransE: 100.

\begin{table*}[]
\centering
\small
\setlength{\tabcolsep}{3.5pt}
\begin{tabular}{c  l  ll  ll   ll  ll }
\toprule
    & & \multicolumn{2}{c}{\textbf{DistMult}} & \multicolumn{2}{c}{\textbf{ComplEx}} & \multicolumn{2}{c}{\textbf{ConvE}} & \multicolumn{2}{c}{\textbf{TransE}} \\
   \cmidrule(lr){3-4}  \cmidrule(lr){5-6}  \cmidrule(lr){7-8} \cmidrule(lr){9-10} 
    & & \textbf{MRR}   & \textbf{Hits@1}  & \textbf{MRR}   & \textbf{Hits@1} & \textbf{MRR}   & \textbf{Hits@1} & \textbf{MRR}   & \textbf{Hits@1} \\
\midrule
     \textbf{Original}  & & 0.82              &  0.67     & 0.99       &   0.99             & 0.80              &   0.63      &    0.65               &  0.45   \\
\midrule
    \multirow{5}{*}{\shortstack[l]{\textbf{Baseline} \\ \textbf{Attacks}}}
    & Random\_n  & 0.80 (-2\%)    &  0.63     & 0.99  (0\%)  &    0.98          & 0.79 (-2\%)       &   0.61       &   0.46 (-29\%)       & 0.18   \\
    & Random\_g1 & 0.82            &  0.66     & 0.99           &   0.98           & 0.80              &   0.62       &    0.57             &  0.33   \\
    & Random\_g2  & 0.81           &  0.65     & 0.99           &   0.98           & 0.79              &   0.62        &  0.50             &  0.22   \\
\cline{2-10} \\[-7pt]
    & Zhang et al.   & 0.77 (-6\%)     &  0.59     & 0.97 (-3\%)    &  0.95            & 0.77  (-3\%)            &  0.61       &  0.43 (-33\%)       &  0.16  \\
    & CRIAGE     & 0.78     &  0.61     &  -              &  -               & 0.78        &  0.63       & -                     &  -   \\
\midrule
    \multirow{9}{*}{\shortstack[l]{\textbf{Proposed} \\ \textbf{Attacks}}}
    & Sym\_truth &  0.62            & 0.30       &  0.90         &   0.82             & \textbf{0.58} (\textbf{-17\%})       &   \textbf{0.27}      & 0.74          &  0.60  \\
    & Sym\_rank  &  0.59            &  0.27      &  0.89 (-10\%)        & 0.79      &  0.62          &  0.33        & 0.52       &   0.34 \\
    & Sym\_cos   &  \textbf{0.50} (\textbf{-38\%}) &  \textbf{0.17} &  0.92   &  0.85   & 0.60            &   0.35        & \textbf{0.41 (-37\%)}         &  \textbf{0.13}  \\
\cline{2-10} \\[-7pt]  
    & Inv\_truth  & 0.81            & 0.66       &   0.86          &   0.74        &  0.78 (-3\%)      &  0.61        & 0.59               & 0.34   \\
    & Inv\_rank   & 0.82            &  0.66      &   \textbf{0.84 (-16\%)}   &   \textbf{0.68}         &  0.79      &   0.61       & 0.55       &  0.34  \\
    & Inv\_cos    & 0.79 (-3\%)      &  0.64     &  0.87   &   0.75        &  0.80              &   0.63        & 0.51 (-22\%)             &  0.25   \\
\cline{2-10} \\[-7pt]
    
    & Com\_truth  & 0.79          &  0.62      &  0.98             &   0.97       &   0.77            &    0.62       & 0.53 (-18\%)               &  0.25  \\
    & Com\_rank   & 0.80          & 0.64      & 0.98               &   0.96       &   0.75 (-6\%)     &   0.58        & 0.67                &  0.47   \\
    & Com\_cos    & 0.78 (-5\%)      &  0.61     & 0.97 (-2\%)      &   0.95      &    0.77            &   0.62        & 0.58        &  0.32   \\
    
\bottomrule    

\end{tabular}
\caption{\small Reduction in MRR and Hits@1 due to different attacks on the \textbf{target split} of WN18. For each block of rows, we report the \emph{best} relative percentage difference from original MRR; computed as $(original-poisoned)/original*100$. Lower values indicate better results; best results for each model are in bold.}
\label{tab:benchmark_mrr_WN18}
\end{table*}

\section{Analysis on Decoy Triples}
\label{apx:decoy_analysis}

The proposed attacks are designed to generate adversarial triples that improve the KGE model performance on decoy triples $(s,\mathtt{r},o')$ and $(s',\mathtt{r},o)$. 
In this section, we analyze whether the performance of KGE models improves or degrades over decoy triples after poisoning.
For the decoy triples on object side $(s,\mathtt{r},o')$, we compute the change in object side MRR relative to the original object side MRR of these triples. Similarly, for the decoy triples on subject side $(s',\mathtt{r},o)$, we compute the change in subject side MRR relative to the original subject side MRR of these decoy triples. Figure \ref{fig:decoy_graphs} shows plots for the mean change in MRR of object and subject side decoy triples. 

We observed in Section \ref{sec:results} that the composition attacks against TransE on WN18RR improved the performance on target triples instead of degrading it. In Figure \ref{fig:decoy_graphs}, we notice that composition attacks against TransE are effective in improving the ranks of decoy triples on both WN18RR and FB15k-237. This evidence supports the argument made in the main paper - it is likely that the composition attack does not work against TransE for WN18RR because the original dataset does not contain any composition relations; thus adding this pattern improves model's performance on \emph{all} triples instead of just the target triples because of the sensitivity of TransE to composition pattern.

% -----------------------------------------------------------------------------------------------------------
% -----------------------------------------------------------------------------------------------------------

\section{Analysis on WN18}
\label{apx:WN18_analysis}
The inversion attacks identify the relation that the KGE model might have learned as inverse of the target triple's relation. But the benchmark datasets WN18RR and FB15k-237 do not contain inverse relations, and a KGE model trained on these clean datasets would not be vulnerable to inversion attacks. Thus, we perform additional evaluation on the WN18 dataset where triples with inverse relations have not been removed. Table \ref{tab:benchmark_mrr_WN18} shows the results for different adversarial attacks on WN18. 

We see that the symmetry based attack is most effective for DistMult, ConvE and TransE. This indicates the sensitivity of these models to the symmetry pattern even when inverse relations are present in the dataset. For DistMult and ConvE, this is likely due to the symmetric nature of their scoring functions; and for TransE, this is likely because of the translation operation as discussed in Section \ref{sec:results}. On the ComplEx model, we see that though the symmetry attacks are more effective than random baselines, the inversion attacks are the most effective. This indicates that the ComplEx model is most sensitive to the inversion pattern when the input dataset contains inverse relations.

% -------------------------------------------------------------------------------------------------------------
% -------------------------------------------------------------------------------------------------------------

 \section{Analysis of Runtime Efficiency}
 \label{apx:runtime_analysis}
 In this section, we compare the runtime efficiency of the baseline and proposed attacks. Table \ref{tab:runtime} shows the time taken (in seconds) to select the adversarial triples using different attack strategies for all models on WN18 dataset. 
 Similar patterns were observed for attack execution on other datasets. 
 
 \begin{table}[]
\centering
\small
\setlength{\tabcolsep}{3.5pt}
\begin{tabular}{c  l  rrrr  }
\toprule
     & & \textbf{DistMult} & \textbf{ComplEx} & \textbf{ConvE} & \textbf{TransE} \\
   
\midrule
    & Random\_n              &  10.08    & 10.69      & 8.76     & 7.83         \\
    & Random\_g1              &  \textbf{8.28}     &  \textbf{8.16}     & \textbf{7.64}    & \textbf{6.49}      \\
    & Random\_g2              &  16.01     &  15.82     &  18.72     & 13.33      \\
\cline{2-6} \\[-7pt]
    & Zhang et al.            &  94.48     &  \textbf{255.53}   & 666.85   & \textbf{81.96}          \\
    & CRIAGE                 & \textbf{21.77}     &  -     & \textbf{21.96}    & -       \\
\midrule
    & Sym\_truth            & \textbf{19.63}            & 35.40     & \textbf{22.76}    & 31.59        \\
    & Sym\_rank             & 23.47            & \textbf{27.25}    & 25.82    &  25.03               \\
    & Sym\_cos               & 22.52            & 28.62       &  25.69   & \textbf{23.13}        \\
\cline{2-6} \\[-7pt] 
    & Inv\_truth           & \textbf{11.43}           & \textbf{15.69}      & 24.13    & 31.89    \\
    & Inv\_rank           & 15.27           & 18.14   & 30.99    & 21.82        \\
    & Inv\_cos           &  14.96           &  20.47      & \textbf{23.02}    & \textbf{20.63}    \\
\cline{2-6} \\[-7pt]
    
    & Com\_truth       &  2749.60           &  1574.44    & 6069.79      & 470.34        \\
    & Com\_rank       &  \textbf{22.04}           &  \textbf{31.53}     & 37.81        & 20.88                 \\
    & Com\_cos       &  34.78           &  68.06      & \textbf{32.37}    &  \textbf{19.86}                 \\
    
\bottomrule    

\end{tabular}
\caption{Time taken in seconds to generate adversarial triples using baseline and proposed attacks on WN18}
\label{tab:runtime}
\end{table}

For CRIAGE, the reported time does not include the time taken to train the auto-encoder model. Similarly, for soft-truth based composition attacks, the reported time does not include the time taken to pre-compute the clusters.
We observe that the proposed attacks are more efficient than the baseline Zhang et al. attack which requires a combinatorial search over the canidate adversarial triples; and have comparable efficiency to CRIAGE. 
Among the different proposed attacks, composition attacks based on soft-truth score take more time than others because they select the decoy entity by computing the soft-truth score for multiple clusters.

\end{document}